\documentclass[preprint,12pt]{elsarticle}

\usepackage{amssymb}
\usepackage{amsmath}
\usepackage{times}
\usepackage{epsfig}
\usepackage{graphicx}
\usepackage{url}
\usepackage{subfig}
\usepackage{booktabs}
\usepackage{pifont}
\usepackage{multirow}
\usepackage[mathscr]{eucal}
\usepackage{amsfonts}
\usepackage[colorlinks=true,pagebackref=true]{hyperref}
\usepackage[left=2.5cm, right=2.5cm, top=3cm, bottom=3cm]{geometry}

\journal{Infrared Physics \& Technology}

\begin{document}

\begin{frontmatter}



\title{Infrared Image Deturbulence Restoration Using Degradation Parameter-Assisted Wide \& Deep Learning}


\author[1]{Yi Lu}

\author[1]{Yadong Wang}

\author[1]{Xingbo Jiang}


\affiliation[1]{organization={ Image Processing Center},
            addressline={Beihang University}, 
            city={Beijing},
            postcode={102206},
            country={China}}

\affiliation[2]{organization={The State Key Laboratory of Virtual Reality Technology and Systems},
            addressline={Beihang University}, 
            city={Beijing},
            postcode={100191}, 
            country={China}}
            
\affiliation[3]{organization={Beijing Advanced Innovation Center for Biomedical Engineering},
            addressline={Beihang University}, 
            city={Beijing},
            postcode={100083}, 
            country={China}}

\author[1,2,3]{Xiangzhi Bai$^*$}
\ead{jackybxz@buaa.edu.cn}
\cortext[1]{Corresponding author}

%% Abstract
\begin{abstract}
      Infrared images captured under turbulent conditions are often degraded by complex geometric distortions and blurring, which substantially compromise image clarity and subsequent analysis. In this paper, we recast the infrared deturbulence problem as a more general image restoration task and propose a parameter assisted image restoration method that leverages degradation prior information from the turbulent infrared images. Specifically, we propose an ingenious and efficient multi-frame image restoration network (DparNet) with wide \& deep architecture, which integrates degraded images and prior knowledge of degradation to reconstruct images with ideal clarity and stability. The degradation prior is directly learned from degraded images in form of key degradation parameter matrix, with no requirement of any off-site knowledge. The wide \& deep architecture in DparNet enables the learned parameters to directly modulate the final restoring results, boosting spatial \& intensity adaptive image restoration. We demonstrate the proposed method on infrared image deturbulence by constructing a dedicated dataset of 49,744 images to rigorously evaluate its performance under challenging turbulence degradation. To further validate the generality of our approach, a supplementary visible image denoising experiment was also conducted on a larger dataset containing 109,536 images. The experimental results show that our DparNet significantly outperform SoTA methods in restoration performance and network efficiency. More importantly, by utilizing the learned degradation parameters via wide \& deep learning, we can improve the PSNR of image restoration by 0.6$\sim$1.1 dB with less than 2$\%$ increasing in model parameter numbers and computational complexity. Our work suggests that degraded images may hide key information of the degradation process, which can be utilized to boost spatial \& intensity adaptive image restoration.
\end{abstract}

%% Keywords
\begin{keyword}


Infrared Image Deturbulence \sep Degradation Prior \sep Parameter-Assisted Restoration \sep Wide and Deep Network

\end{keyword}
    
\end{frontmatter}



\section{Introduction}

\begin{figure}[!htbp]
	\begin{center}
		\includegraphics[width=0.8\linewidth]{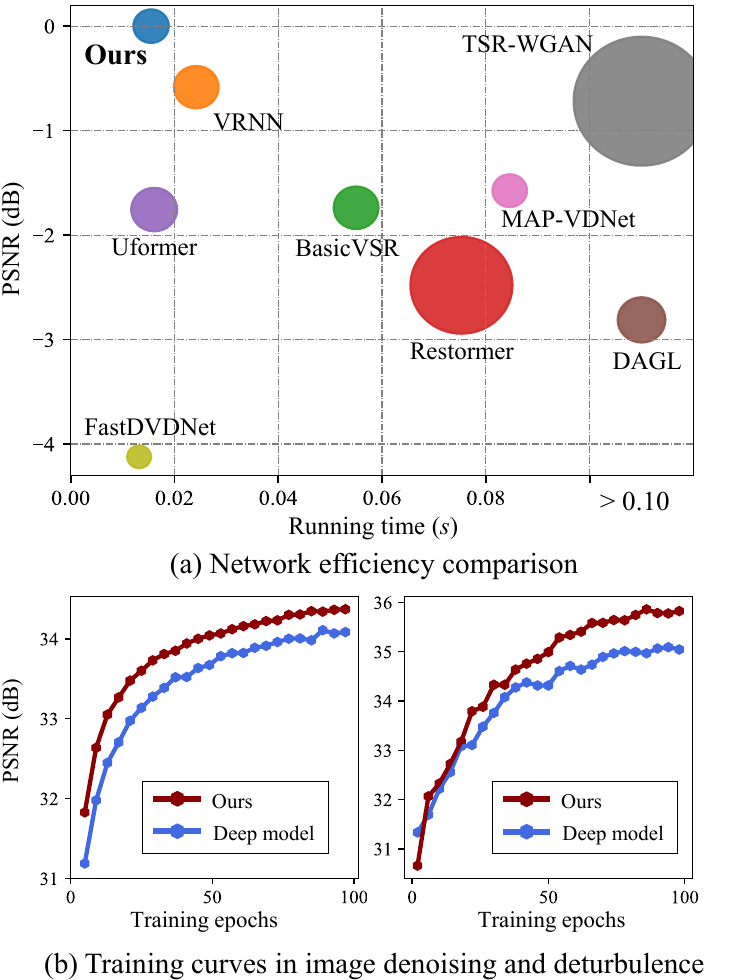}
	\end{center}
	\caption{Wide \& deep learning boosts efficient image restoration. (a) is drawn according to relative PSNR  and average running time to restore a \texorpdfstring{$256 \times 256 \times 3$}{256 x 256 x 3} image for our method and SoTA restoration methods. Area of each circle in (a) is proportional to the number of model parameters. (b) shows the promotion of wide \& deep learning on improving restoration performance.}
	\label{fig.1}
\end{figure}

Infrared imaging suffers from various types of image degradation, which can be caused by the physical limitations of the camera or by harsh imaging environments. For instance, at medium or long distances in hot weather, the disturbance of optical path caused by atmospheric turbulence leads to multiple forms of image degradation, such as geometric distortion and blurring \cite{Lohse2010, Xue2013}. Such degradations severely degrade the quality of image or video, which affects subsequent imaging-based civilian or military applications such as security surveillance and visual navigation guidance.

Numerous attempts have been made to mitigate the negative effects of image degradation. Traditional methods rely on modelling degradation to approximate the inverse process and realise image restoration \cite{Elad2006, Shimizu2008}. 
In recent years, deep learning has freed researchers from complex modelling and optimisation, enabling establishment of end-to-end mapping from degraded image to restored image \cite{Kim2017, Zhou2019, Su2017}. 
However, most existing deep learning-based methods focus only on addressing of image degradation with constant intensity and uniform spatial distribution \cite{tassano2020fastdvdnet, jin2021neutralizing, mou2021dynamic, zamir2022restormer, wang2023versatile}, which does not adequately address the spatially inhomogeneous and intensity-varying degradations commonly encountered in infrared images affected by turbulence.
In addition, most existing methods make insufficient use of degradation prior knowledge when performing restoration, leading to low efficiency and poor interpretability. There lacks methods capable of handling spatially inhomogeneous, intensity-varying degradation, which is of great challenge and research value.

Notably, the degradation information inherent in the imaging process governs the manifestation, spatial distribution, and intensity variation of the degradation, providing a potent source of prior knowledge for image restoration, which is particularly useful for addressing infrared deturbulence. 
The difficulty of obtaining prior information has been puzzling the field of image restoration, forcing most deep learning methods to adopt a data-driven blind restoration strategy. Nevertheless, we assume that the degradation prior is just hidden in degraded images, and it can be learned in a parametrised manner to guide neural networks in image restoration.

In this paper, we recast the infrared deturbulence problem as a general image restoration task and propose an efficient framework to perform intensity \& spatially adaptive multi-frame image restoration. 
Specifically, we first learn key degradation parameters from degraded images through parameter prediction network with a simple encoder-decoder architecture. The learned parameters and degraded sequences are fed together into our degradation parameter assisted restoration network (DparNet in Figure \ref{fig.overall}). The proposed DparNet adopts a wide \& deep architecture, inspired by Google's wide \& deep learning for recommender systems \cite{cheng2016wide}. The deep model performs in-depth down-sampling encoding on the input degraded sequence and produce high quality restored sequence end-to-end. The wide model, with a shallow encoding depth, wide feature resolution and small number of model parameters, integrates both the learned degradation parameters and degraded image sequence. The deep and wide models are implemented in parallel, with final restoring result being a fusion of the outputs from these two sub-models. With the architecture of wide \& deep models, DparNet makes full use of learned degradation parameters to guide the network in suppressing degradation varies in space and intensity, improving restoration performance remarkably with negligible increase in model size and computational complexity. Notably, by leveraging a parameter prediction network to extract degradation prior directly from degraded images, proposed method is theoretically extendable to other imaging modalities and degradation types.

We demonstrate the proposed method on the infrared image deturbulence task by constructing a dedicated dataset of 49,744 images to rigorously evaluate its performance under challenging turbulence degradation, and further validate its generality through supplementary visible image denoising experiments on a larger dataset of 109,536 images. Experimental results show that our DparNet can effectively suppress degradation with spatial and intensity variations, and significantly outperform SoTA restoration methods. In addition, our method has been rationally and ingeniously designed to achieve particularly high network efficiency, i.e., achieving excellent restoration performance with high processing speed, as shown in Figure \ref{fig.1} (a). More importantly, utilization of degradation prior by wide \& deep learning greatly enhances restoration performance, as shown in training curves illustrated by Figure \ref{fig.1} (b). Quantitative results of the ablation study show that PSNR for image denoising and deturbulence is improved by 0.6 and 1.1 dB respectively via wide \& deep learning, while the model size and computational complexity increase by less than 2\%.

Overall, the main contributions can be summarised as: 
\begin{enumerate}
	\item New insight of learning degradation prior from degraded images to promote image restoration is explored, filling the gap in research of spatial \& intensity adaptive image restoration, addressing the challenging problem of infrared deturbulence.
	\item We propose a novel multi-frame image restoration network (DparNet), which exploits learned degradation parameters to efficiently boost image restoration through a wide \& deep architecture.
	\item We construct dedicated large-scale datasets for infrared image deturbulence (49,744 images) and visible image denoising (109,536 images), facilitating rigorous evaluation and comparison with state-of-the-art methods.
	\item Extensive experiments demonstrate the promotion to restoration performance from utilization of degradation prior, and our DparNet outperforms SoTA methods in restoration performance and network efficiency.
\end{enumerate}

\section{Related work}
\subsection{Image restoration}
Image restoration, an important fundamental technique in computer vision, aims to use single or multiple degraded images with prior knowledge to obtain the desired image \cite{banham1997digital}. Compared with upgrading imaging hardware to avoid degradation \cite{Rigaut2018}, restoring ideal image from degraded image is a more cost-effective solution. Traditional image restoration methods model the degradation process and then iterate through mathematical optimizations to approximate the ideal image \cite{Elad2006, Dong2012}. Accurate modelling the degradation is crucial for restoring ideal images, which limits the application value of traditional restoration methods. Deep learning has enabled the construction of end-to-end mappings from degraded image to restored image. In recent years, deep learning-based image restoration methods have demonstrated outstanding performance \cite{jin2021neutralizing, mou2021dynamic, zamir2022restormer, wang2023versatile}. However, most existing deep learning-based methods concentrate on degradation with constant intensity and uniform spatial distribution, such as random noise with fixed intensity level \cite{wang2023versatile, sun2021deep, vaksman2021patch}. Although one research attempt to locate the spatial distribution of degradation to guide spatially adaptive image restoration \cite{purohit2021spatially}, it still overlooks changes in degradation intensity and does not fully utilize available degradation prior information. In this paper, we aim to fully exploit the degradation prior concealed in degraded images to suppress degradation that vary in both space and intensity.

\subsection{Image denoising}
Image denoising is a classic image restoration task with a extensive research history. An early image denoising technique integrates similar pixel information from neighboring regions through non-linear filtering to reduce image noise \cite{Smith1997}. Subsequent traditional methods employed sparse representations or regional self-similarity to improve denoising performance \cite{Elad2006, Dong2012, Dabov2007}. More recently, deep learning has greatly promoted the performance of image denoising \cite{Guo2019, tassano2020fastdvdnet, sun2021deep}. Since end-to-end methods rely on large amounts of training data rather than degradation modelling, some recent research has been devoted to developing general frameworks capable of handling various types of image degradation, and these methods have become SoTA methods for image denoising \cite{zamir2022restormer, Wang_2022_CVPR, wang2023versatile}. However, these deep learning-based methods are content to suppress spatially uniform noise with constant intensity level and lack the ability to handle complex noise varies in intensity and spatial distribution.

\subsection{Image deturbulence}
In contrast to image degradation due to physical limitations of camera hardware, the atmospheric turbulence degradation is predominantly caused by the inherent physical properties of the imaging medium \cite{jin2021neutralizing}. Atmospheric turbulence causes irregular fluctuations in refractive index of atmosphere, leading to bending and dissipation of optical path, which results in strong random geometric distortions and blurring in the captured images \cite{Megaw1950, Lohse2010, Datcu2005}. To mitigate atmospheric turbulence degradation in infrared or visible light images, traditional methods adopt multi-frame registration-fusion and deconvolution techniques to realize image deturbulence \cite{Hirsch2010, Anantrasirichai2013, Chan2011, Lou2013}. However, the difficulty of collecting paired training data has hindered the development of deep learning-based methods for image deturbulence. Recently, one study overcame this challenge by constructing datasets using algorithm and heat source simulations, achieving promising results in image deturbulence \cite{jin2021neutralizing}.

\section{The proposed method}
\begin{figure*}[!htbp]
	\begin{center}
		\includegraphics[width=0.9\linewidth]{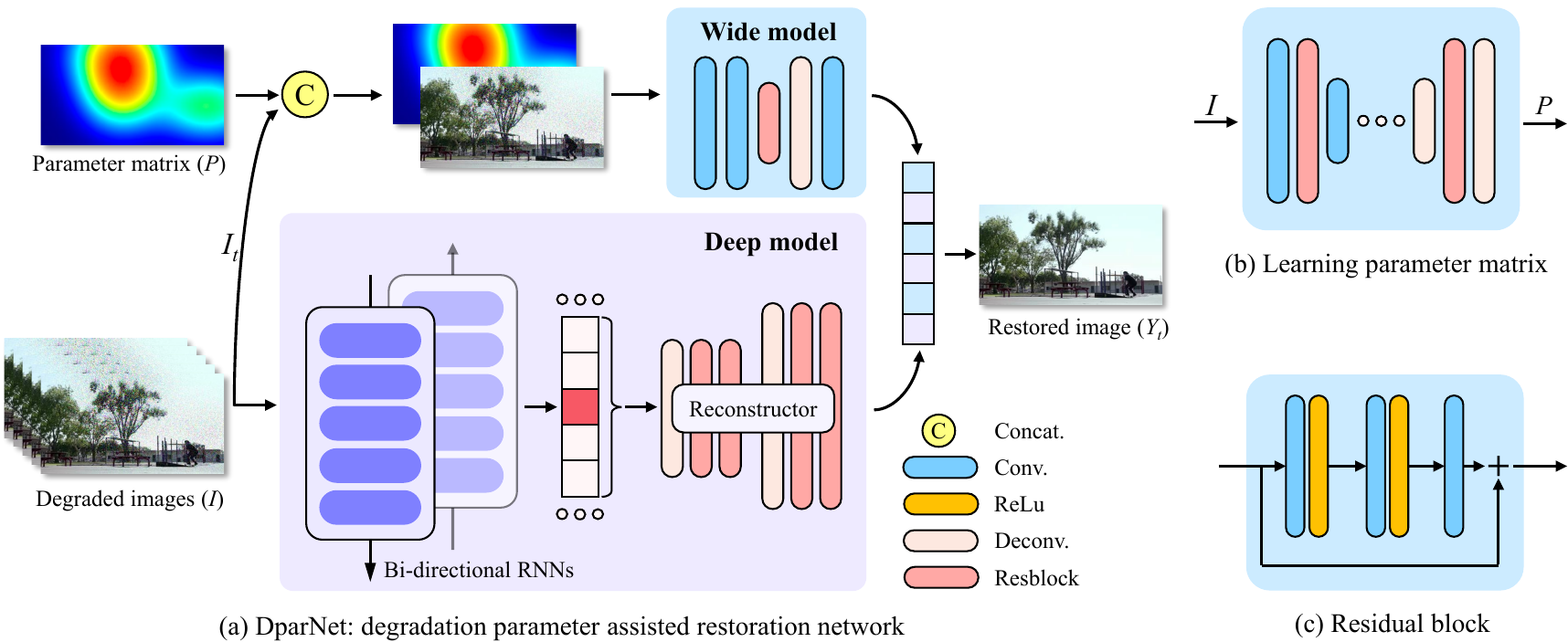}
	\end{center}
	\caption{Overall structure of the proposed method. The learned degradation parameter matrix ({\itshape P}) is integrated with degraded images ({\itshape I}) via our DparNet in (a), which adopts a wide \& deep architecture, to generate restored images ({\itshape Y}).}
	\label{fig.overall}
\end{figure*}

\subsection{Problem formulation and motivation}

In general, the degradation process of an image can be formulated as:
\begin{equation}
	D = H(C),
\end{equation}
where $D$ and $C$ denote the degraded and clean images, respectively, and $H(\cdot)$ represents the degradation function. When using deep neural network (DNN) to solve the inverse process of formula (1), the optimization of DNN can be formulated as \cite{wang2023versatile}:
\begin{subequations}
	\begin{align}
		&W^* = \mathop{\arg\min}_{W} \| {\rm DNN}(D; W) - C \|,\\
		&\hat{C} = {\rm DNN}(D; W^*),
	\end{align}
\end{subequations}
where ${\rm DNN}(\cdot; W)$ represents a deep neural network with parameter $W$. The trained DNN can approximate $H^{-1}$ to achieve image restoration. 

However, when both the intensity and spatial distribution of degradation vary, the effectiveness of a trained DNN diminishes. The reason is that a fixed DNN cannot approximate the reverse process of a dynamically changing $H$.
Considering the variability of degradation, formula (1) should be modified to:
\begin{equation}
	D = H(C;P),
\end{equation}
where $P$ denotes key degradation parameters that characterise the degradation. $P$ represents different parameters in different types of degradation. For instance, $P$ may include turbulence intensity $C_n^2$ and noise intensity level $\sigma_n$ in atmospheric turbulence degradation and noise degradation, respectively. In fact, $P$ could be any parameter that dominates degradation. Considering the spatial distribution of degradation, $P$ is supposed to be a two-dimensional matrix to match the spatial distribution of the degraded image. Based on formula (3), an straightforward improvement to formula (2) is to replace ${\rm DNN}(D;W)$ with ${\rm DNN}(D,P;W)$, as follows:
\begin{subequations}
	\begin{align}
		&W^* = \mathop{\arg\min}_{W} \| {\rm DNN}(D, P; W) - C \|,\\
		&\hat{C} = {\rm DNN}(D, P; W^*).
	\end{align}
\end{subequations}
Although the improvement in formula (4) could be helpful, incorporating the degradation prior as an additional input may cause it to be overwhelmed by dense image features. \cite{cheng2016wide}. In fact, a simpler architecture can allow the prior knowledge to directly influence the final result \cite{sheng2021one}. 
Introducing degradation prior through simple architecture is a more potential strategy. Based on this motivation, we give the restoration paradigm of this paper:
\begin{equation}
\hat{C} = {\rm Merge}({\rm DM}(D;W_D), {\rm WM}(D,P;W_W)),
\end{equation} 
where ${\rm DM}$ denote a deep model, i.e., DNN mentioned above, ${\rm WM}$ is a wide model with simple architecture and wide feature resolution. Through the concise and effective paradigm in formula (5), we not only retain the powerful non-linear mapping ability of the deep model, but also allows prior knowledge to modulate the final restoration result without hindrance. 

An important question remains: how to obtain the parameter matrix $P$. Notably, since the manifestation, spatial distribution and intensity variation of degradation are highly related to characteristics of key degradation parameters, the possibility of revealing degradation parameter matrix directly from degraded images exists, which has potential to provide an easily-available and powerful prior for image restoration. Based on the above and motivations, we have developed the framework of this paper.

\subsection{Overall workflow}
We propose an efficient and ingenious framework for spatial \& intensity adaptive image restoration (see Figure \ref{fig.overall}), whose overall workflow can be outlined in four steps. Firstly, a parameter prediction network consisting of several convolutional and deconvolutional layers, is employed to learn the numerical matrix of key degradation parameters, such as noise intensity level $\sigma_n$ or turbulence intensity $C_n^2$, from the input degraded image sequence. Secondly, the degraded images and learned parameter matrix are fed into the proposed degradation parameter assisted restoration network (DparNet) for parallel processing. The wide \& deep architecture in DparNet enables learned parameter matrix to directly influence the reconstruction results as a degradation prior rather than being submerged within dense image features. Thirdly, the outputs of wide \& deep models are merged to obtain the final restoration result. Finally, during the test phase, the only required input for entire framework is the degraded image sequence, while output contains both degradation parameter matrix and restored image sequence.

\subsection{The proposed DparNet}
The Architecture of the proposed DparNet is illustrated in Figure \ref{fig.overall}(a). DparNet adopts a wide \& deep architecture. The deep model is an efficient multi-frame image restoration network that enables end-to-end mapping from degraded sequence to restored sequence. The deep model performs multi-layer downsampling and deep encoding of the degraded sequence, and then reconstructs restored sequence from the extracted image features. Specifically, the input degraded sequence ($I$) is first encoded into dense image features via a feature extractor implemented using a bi-directional recurrent neural network (BRNN). The BRNN we used, streamlined from the version proposed by Wang et al., has been demonstrated to be effective for wide types of multi-frame image restoration \cite{wang2023versatile}. We streamlined the BRNN from Wang et al. by reducing the number of residual dense blocks by half \cite{zhang2020residual} to achieve high network efficiency. The extracted features, denoted as $F$, are decoded by a reconstructor (RC) composed of several residual blocks and transposed convolutional layers. Each restored target frame ($Y_t$) is reconstructed by fusing its feature map and the feature maps of four neighbouring frames. The workflow of deep model can be formulated as follows:
\begin{subequations}
	\begin{align}
		\dots, F_t, F_{t+1}, \dots={\rm BRNN}(\dots, I_{t}, I_{t+1}, \dots),& \\
		Y_t={\rm RC}({\rm Cat}(F_{t-2}, F_{t-1}, F_t, F_{t+1}, F_{t+2})),&
	\end{align}
\end{subequations}
where $\rm Cat(\cdot)$ is the operation of concatenation.

Additionally, the degraded target frame, concatenated with the learned parameter matrix, is fed into the wide model. Our wide model containing only three convolutional layers, one deconvolutional layer and one residual block, as shown in Figure \ref{fig.overall} (a). The number of parameters in the wide model is approximately 1\% of that in the deep model, which aims to introduce the assistance of degradation prior without incurring a perceptible increase in computational burden.
The image features maintain a ``wide'' resolution in wide model, different from deep downsampling and encoding in deep model. The shallow encoding level and wide resolution in wide model enable parameter matrix to directly modulate the final result. The restored image is obtained by merging the outputs of the wide \& deep model, as follows:
\begin{subequations}
	\begin{align}
		\dots, Y_t^1, &Y_{t+1}^1, \dots={\rm DM}(\dots, I_{t}, I_{t+1}, \dots), \\
		Y_t^2&={\rm WM}({\rm Cat}(I_{t}, P)), \\
		Y_t&={\rm Conv}_{1\times1}({\rm Cat}(Y_t^1, Y_t^2)),
	\end{align}
\end{subequations}
where $\rm DM(\cdot)$ and $\rm WM(\cdot)$ denote the processing by deep and wide models, ${\rm Conv}_{1\times1}(\cdot)$ means adjusting the channels via a convolutional layer with kernel size of $1\times1$. Each reconstructed frame is then concatenated in temporal order to obtain the restored sequence.

Overall, our framework is developed based on a simple but non-trivial insight:  degradation information hidden in degraded images can be revealed and used as prior knowledge to facilitate image restoration. In order to avoid the degraded parameter matrix being submerged by dense image features extracted by deep model, we adopt a wide \& deep architecture to make parameter matrix directly affect the final restoration result. Based on the above designs, our DparNet has the potential to achieve efficient spatial \& intensity adaptive image restoration.

\subsection{Implementation details}
Our DparNet is trained using a loss function composed of pixel loss and perceptual loss \cite{ledig2017photo}, as follows:
\begin{equation}
	L=\alpha_1L_{pixel}+\alpha_2L_{perceptual},
\end{equation}
where $L$ denotes the loss function, $\alpha_1$ and $\alpha_2$ are taken as 1 and 0.05 respectively in this work. The pixel loss was calculated through the $\mathcal{L}_1$ distance between restored image and ground truth. The perceptual loss was obtained by feeding the restored image and ground truth into a pre-trained feature extraction model (VGG19 \cite{simonyan2014very}) and calculating $\mathcal{L}_1$ distance between the outputs. In addition, the parameter prediction network was trained via the $\mathcal{L}_1$ distance between the predicted parameter matrix and ground truth.

During training, we randomly cropped input image into $256\times256$ patches and applied random horizontal and/or vertical flips. The Adam \cite{kingma2014adam} optimizer was employed to train our model for 100 epochs with a learning rate of 0.0001. Our experiments were conducted on a platform with Windows 10 system and two NVIDIA RTX 3090-Ti graphics cards.

\section{Experimental results}
\subsection{Dataset construction and evaluation manners}
We conducted experiments on two representative restoration tasks, with a primary focus on infrared image deturbulence and supplementary evaluation on image denoising. Datasets for these two tasks, with degradation changes in space and intensity, were constructed for training and evaluating the proposed method and comparison methods. The 
deturbulence dataset was constructed through outdoor photography using a FLIR A615 thermal infrared camera. We simulated the atmospheric turbulence degradation (geometric distortion and blurring) with complex random spatial distribution and random intensity ($0 \leq C_n^2 \leq 6\times10^{-12}$) on clear images using  simulation algorithm from Jin et al. \cite{jin2021neutralizing}. The training set contains 1,302 pairs of degraded-clean sequences with a total of 39,060 images. The test set contains 40 pairs of degraded-clean sequences with a total of 10,684 images. The training sequence has been standardised to 15 frames, while the frame length of test sequence varies from tens to hundreds. The resolution of our deturbulence data is $480\times640$.

Another aspect, our denoising dataset was constructed on the basis of Vimeo-90K dataset \cite{Xue2019}. Additive random noise with complex random spatial distribution and random intensity level ($0 \leq \sigma_n \leq 100$) was added to clear images to obtain noise-degraded images as network inputs. The denoising dataset contains 7,824 pairs of degraded-clean sequences with a total of 109,536 images with resolution of $256\times448\times3$. The frame length of denoising dataset if 7, according to the pre-division of Vimeo-90K dataset. The 20\% of data was used for test, and the rest was used for training and validation.

Quantitative and qualitative assessments have been conducted to evaluate the restoration performance. Adopted quantitative metrics include Peak Signal to Noise Ratio (PSNR), Structural Similarity (SSIM) \cite{Wang2004}, Normalized Root Mean Square Error (NRMSE), and Variation of Information (VI) \cite{Meila2007}. Qualitative assessments are conducted through the visualisation of restoring results.
In addition to evaluating the restoration performance, we compare network efficiency of the proposed model and comparison models by counting the number of model parameters, the average FLOPs and running time.
All the comparison methods adopted the same data augmentation and training strategies as the proposed method.

\subsection{Performance on image deturbulence}
\begin{table}
	\footnotesize
	\begin{center}
		\begin{tabular}{l|cccc}
			\hline
			Methods & PSNR ($\uparrow$) &SSIM ($\uparrow$) & NRMSE ($\downarrow$) & VI ($\downarrow$)\\
			\hline
			SGF \cite{Lou2013}&27.8677&0.8562  &0.0766&8.8624  \\
			CLEAR \cite{Anantrasirichai2013} &32.2483&0.8999  &0.0475&7.5707  \\
			BasicVSR \cite{Chan2021} &31.1472&0.8719   &0.0553&7.3905   \\
			TSR-WGAN \cite{jin2021neutralizing} &33.2667&0.9065  &0.0427&7.0831   \\
			DAGL \cite{mou2021dynamic}&30.4874&0.8427 &0.0583&8.1927 \\
			Restormer \cite{zamir2022restormer}&30.1516&0.8384 &0.0599&8.1288  \\
			VRNN \cite{wang2023versatile} &33.3376&0.9081  &0.0421&7.0141  \\
			Ours & {\bfseries 33.9853}&{\bfseries0.9195} & {\bfseries 0.0397}&{\bfseries 6.7744} \\
			\hline
		\end{tabular}
	\end{center}
	\caption{Quantitative comparison of image deturbulence.}
	\label{tab.deturbulence}
\end{table}

\begin{figure*}
	\begin{center}
		\includegraphics[width=0.9\linewidth]{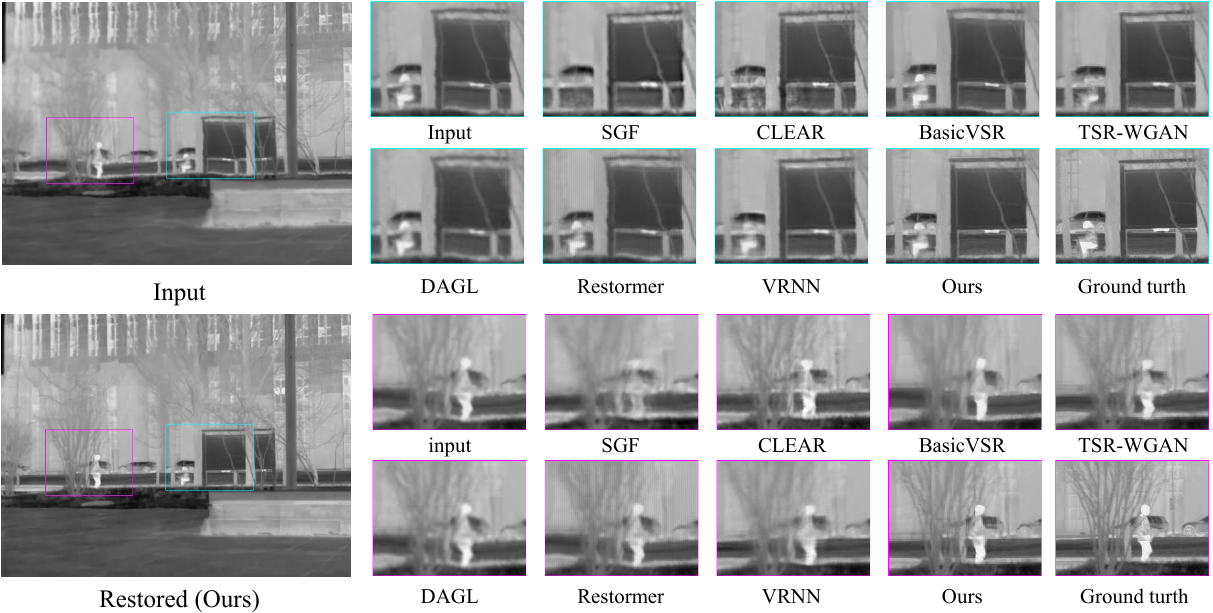}
	\end{center}
	\caption{Visual comparison of image deturbulence.}
	\label{fig.deturbulence}
\end{figure*}
\begin{figure}
	\begin{center}
		\includegraphics[width=0.9\linewidth]{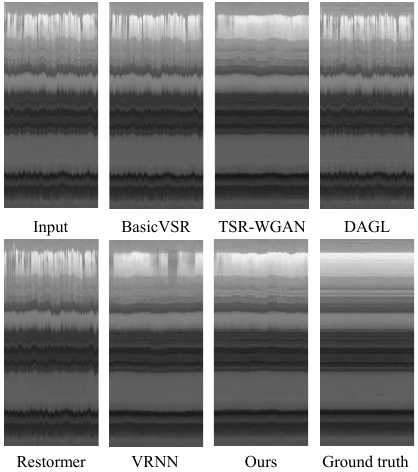}
	\end{center}
	\caption{Stability comparison of image sequences.}
	\label{fig.deturbulence2}
\end{figure}

We compare the proposed DparNet with 7 image/video restoration or deturbulence methods, including 2 traditional registration-fusion-based methods: SGF \cite{Lou2013} and CLEAR \cite{Anantrasirichai2013}, along with 5 SoTA deep learning-based methods: BasicVSR \cite{Chan2021}, TSR-WGAN \cite{jin2021neutralizing}, DAGL \cite{mou2021dynamic}, Restormer \cite{zamir2022restormer}, and VRNN \cite{wang2023versatile}. Quantitative metrics are reported in Table \ref{tab.deturbulence}. The proposed method achieves the best results in all four image quality assessment metrics, with PSNR of our method being 0.65 to 6.12 dB higher than that of comparison methods. Another aspect, from the subjective visual comparison shown in Figure \ref{fig.deturbulence}, the proposed method effectively suppresses the spatial \& intensity varying turbulence degradation in the input image, including geometric distortion and blurring, obtaining the clearest, closest-to-ground truth restoring results. In the first comparison with relatively weak degradation intensity, VRNN, TSR-WGAN and our method achieve acceptable restoration results, among which our result is the best. Advantages of the proposed method become more apparent when the intensity of degradation increases, as shown in the second comparison of \ref{fig.deturbulence}. Another aspect, although traditional deturbulence methods can restore the static background, they destroy the moving target (see SGF and CLEAR restoring results for pedestrian and cyclist). Further, we visualise the stability of sequences by arranging one column of pixels in temporal order. The comparison for deep learning-based methods are shown in Figure \ref{fig.deturbulence2}. The proposed method effectively suppresses the random geometric distortion in input sequence and obtains the most stable restoration results against comparison methods. The above results show that the proposed method outperforms SoTA methods objectively and subjectively in image deturbulence. Notably, since the turbulence degradation is more complex than other dedegradation problems, the great advantages of the proposed method in image deturbulence show its great potential to settle complex degradation.

\subsection{Performance on image denoising}

\begin{table}[!htbp]
	\footnotesize
	\begin{center}
		\begin{tabular}{l|cccc}
			\hline
			Methods & PSNR ($\uparrow$) &SSIM ($\uparrow$) & NRMSE ($\downarrow$) & VI ($\downarrow$)\\
			\hline
			FastDVDNet \cite{tassano2020fastdvdnet} &29.2781&0.8169  &0.1821&9.3020 \\
			MAP-VDNet \cite{sun2021deep} &31.8261&0.8899  &0.0770&8.1362 \\
			BasicVSR \cite{Chan2021} &32.9615&0.9028  &0.0693&7.9442 \\
			DAGL \cite{mou2021dynamic} &31.2747&0.8815  &0.0820&8.3634  \\
			Uformer \cite{Wang_2022_CVPR} &31.6437&0.8792 &0.0799&8.2328 \\
			Restormer \cite{zamir2022restormer} &32.2736&0.8926  &0.0736&7.9846  \\
			VRNN \cite{wang2023versatile} &32.8921&0.9052  &0.0691&7.7636  \\
			Ours & {\bfseries 33.4012} &{\bfseries 0.9143} & {\bfseries 0.0652} & {\bfseries7.7184} \\
			\hline
		\end{tabular}
	\end{center}
	\caption{Quantitative comparison of image denoising. Symbol ($\uparrow$) indicates that higher values have better performance, while ($\downarrow$) is the opposite. The best results are {\bfseries in bold.}}
	\label{tab.denoising}
\end{table}

\begin{figure*}[!htbp]
	\begin{center}
		\includegraphics[width=0.9\linewidth]{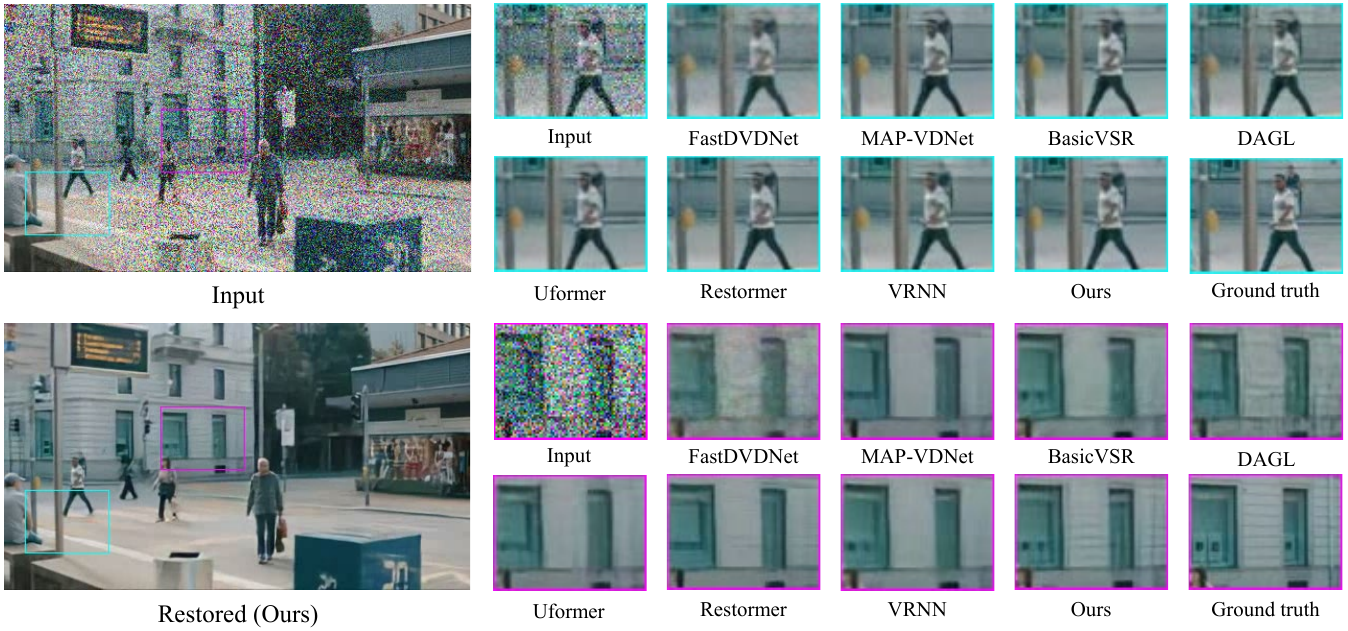}
	\end{center}
	\caption{Visual comparison of image denoising.}
	\label{fig.denoising}
\end{figure*}

As a supplementary evaluation to demonstrate the generalization of our method, we compare the proposed DparNet with 7 SoTA image/video denoising or restoration methods based on deep learning, including FastDVDNet \cite{tassano2020fastdvdnet}, MAP-VDNet \cite{sun2021deep}, BasicVSR \cite{Chan2021}, DAGL \cite{mou2021dynamic}, Uformer \cite{Wang_2022_CVPR}, Restormer \cite{zamir2022restormer}, and VRNN \cite{wang2023versatile}. Quantitative metrics for evaluating restoration performance of the proposed method and comparison methods are reported in Table \ref{tab.denoising}. The proposed method achieves the best results in all four image quality assessment metrics, with PSNR of our method being 0.44 to 4.12 dB higher than that of the comparison methods. From the subjective visual comparison shown in Figure \ref{fig.denoising}, the proposed method effectively suppresses the spatial \& intensity varying noise in input image and achieves the restoration results closest to ground truth. For areas with weak noise intensity, the restoration results of most comparison methods are acceptable, such as the restoration of pedestrian objects in Figure \ref{fig.denoising}. However, when noise intensity increases, the restoration performance of comparison methods decreases. In the second comparison of Figure \ref{fig.denoising}, almost all comparison methods fail to restore textures of the building, only our method can suppress strong noise and restore image details well. The above results show that the proposed method outperforms SoTA methods objectively and subjectively in image denoising.

\begin{table}
	\footnotesize
	\begin{center}
		\begin{tabular}{l|cccc}
			\hline
			Models & Params & FLOPs & Time ({\itshape s})& PSNR\\
			\hline
			Ours &3.1496  &3.4924  &0.0155 &0  \\
			VRNN \cite{wang2023versatile} &5.0357  &4.6890 &0.0242 &$-$0.5839 \\
			TSR-WGAN \cite{jin2021neutralizing} &46.2849  &31.2422  &0.1793 &$-$0.7186 \\
			MAP-VDNet \cite{sun2021deep}&3.0142  &29.1712 &0.0846 &$-$1.5751   \\
			BasicVSR \cite{Chan2021}&4.0756  &27.6860  &0.0550 &$-$1.7389 \\
			Uformer \cite{Wang_2022_CVPR}&5.2919 &1.0684 &0.0161 &$-$1.7563 \\
			Restormer \cite{zamir2022restormer}&26.1266  &14.0990 &0.0753 &$-$2.4807\\
			DAGL \cite{mou2021dynamic}&5.7297  &27.3386 &0.5449 &$-$2.8122 \\
			FastDVDNet \cite{tassano2020fastdvdnet}&1.4598  &4.3679  &0.0132 & $-$4.1231  \\
			\hline
		\end{tabular}
	\end{center}
	\caption{Comparison of network efficiency. Params are number of model parameters divided by $10^6$, FLOPs are average floating point operations, divided by $10^{10}$, to process a frame with shape of $256\times256\times3$. Time ({\itshape s}) is the average running time for a model to process a frame, with shape of $256\times256\times3$, in 100 testing rounds on a RTX 3090-Ti graphics card. If a model has been used in multiple tasks, the average indicator is reported.}
	\label{tab.effieiency}
\end{table}

\subsection{Network efficiency analysis}
Thanks to its rational and innovative structural design, our proposed method is highly efficient, achieving excellent restoration performance with low model complexity. Table \ref{tab.effieiency} reports the comparison of network efficiency between the proposed method and all deep learning-based comparison methods. Model parameters, average floating point operations (FLOPs), running time and relative PSNR are counted. It can be seen that the proposed model has the best restoration performance. 
Meanwhile, our model outperform most comparison models in model parameters, FLOPs, and running time, only slightly inferior than some methods whose restoration performance is noticeably weaker than our model, such as FastDVDNet. Figure \ref{fig.1} (a), plotted based on model parameters, running times and relative PSNR, visually demonstrates the advantages of our method in network efficiency.

\begin{table*}[t]
  \footnotesize
  \begin{center}
    \begin{tabular}{ccc|ccc}
      \hline
      Model & $P$ & W\&D & Params & FLOPs & Time ({\itshape s}) \\
      \hline
      Variant-1 & \ding{55} & \ding{55} & 3.1074 & 3.4302 & 0.0152 \\
      Variant-2 & \ding{51} & \ding{55} & 3.1082 & 3.4399 & 0.0153 \\
      Variant-3 & \ding{55} & \ding{51} & 3.1496 & 3.4924 & 0.0155 \\
      Ours      & \ding{51} & \ding{51} & 3.1496 & 3.4924 & 0.0155 \\
      \hline
    \end{tabular}
    
    \bigskip

    \begin{tabular}{l|cccc}
      \hline
      Model & PSNR ($\uparrow$) & SSIM ($\uparrow$) & NRMSE ($\downarrow$) & VI ($\downarrow$)\\
      \hline
      Variant-1 & 32.8944/32.7923 & 0.9121/0.9082 & 0.0442/0.0675 & 6.9491/7.8012 \\
      Variant-2 & 33.0195/32.7881 & 0.9050/0.8996 & 0.0438/0.0706 & 7.1154/7.8024 \\
      Variant-3 & 33.3579/32.8699 & 0.9101/0.9120 & 0.0424/0.0702 & 6.9992/7.8702 \\
      Ours      & 33.9853/33.4012 & 0.9195/0.9143 & 0.0397/0.0652 & 6.7744/7.7184 \\
      \hline
    \end{tabular}
  \end{center}
  \caption{Results of ablation study. $P$ denotes the degradation parameter matrix, ``W\&D'' denotes wide \& deep learning, ``A/B'' indicates that the metrics for deturbulence and denoising are A and B, respectively.}
  \label{tab.ablation}
\end{table*}

\subsection{Ablation study}
In ablation study, we validate the effectiveness of wide \& deep learning and utilization of degradation prior to boost image restoration, as shown in Table \ref{tab.ablation}. We have made several variants of the proposed method. Variant-1 indicates using our deep model without parameter matrix assistance. Variant-2 indicates using deep model with parameter matrix as an additional input channel. Variant-3 uses the same wide \& deep models as ours, but the parameter matrix is  replaced by a matrix in which all elements are equal to 1. 

The results of variant-2 indicate that directly using learned parameter matrix as additional input into the deep model does not result in a significant performance improvement, which coincides with our assumption. Meanwhile, results of variant-3 show that using wide \& deep architecture, without the assistance of degradation parameter matrix, has improvement in some restoration metrics, such as PSNR. However, gains of two variants mentioned above are weak compared to the facilitation brought by utilizing parameter matrix through wide \& deep learning. Our method improves PSNR by 0.61 and 1.09 dB in image denoising and deturbulence, respectively, with less than $2\%$ increasing in model parameter numbers and computational complexity. Further, the training curves in Figure \ref{fig.1} (b) clearly show that utilizing parameter matrix through wide \& deep learning significantly improve restoration performance. Notably, degradation parameter assistance are more powerful in the task of image deturbulence than in denoising. Given the diverse and stochastic nature of atmospheric turbulence degradation, this preference suggests that our framework is particularly competent  to settle complex image degradation.

\section{Conclusions}
In this paper, we propose a SOTA method for infrared image deturbulence by harnessing degradation priors parameters through a novel wide \& deep architecture. Proposed method is designed specifically to address the spatially and intensity-varying degradations inherent in turbulent infrared imaging. Extensive experiments on a dedicated infrared deturbulence dataset demonstrate that our approach achieves superior restoration performance with remarkably low computational burden. Furthermore, supplementary experiments on image denoising confirm the generalizability and robustness of the proposed framework. Overall, our work not only bridges the gap in spatial and intensity adaptive image restoration but also paves the way for efficient multi-frame restoration in complex imaging scenarios.

\section*{Conflict of interest}
There is no conflict of interest.

\section*{Acknowledgements}  

This work is supported in part by the National Natural Science Foundation of China under Grant 62271016, in part by the Beijing Natural Science Foundation under Grant 4222007, and in part by the Fundamental Research Funds for the Central Universities.

\section*{Data availability}

The data supporting the results of this paper have not been publicly released, but they are available from the authors upon reasonable request.

\bibliographystyle{elsarticle-num}
\bibliography{egbib}

\end{document}